\documentclass[11pt,a4paper]{article}
\usepackage[hyperref]{acl2020}
\usepackage{times}
\usepackage{latexsym}

\usepackage{xcolor}
\usepackage{graphicx}
\usepackage{booktabs}
\usepackage{mathtools}
\usepackage{bbm}
\usepackage{enumitem}
\usepackage{multirow}
\usepackage[explicit]{titlesec}

\newcommand{\todo}[1]{\textbf{\textcolor{red}{TODO: #1}}}
\newcommand{\sr}[1]{\textbf{\textcolor{orange}{[SR: #1]}}}
\newcommand{\gn}[1]{\textbf{\textcolor{blue}{[GN: #1]}}}
\newcommand{\sz}[1]{\textbf{\textcolor{red}{[SZ: #1]}}}
\renewcommand{\todo}[1]{}
\renewcommand{\sr}[1]{}
\renewcommand{\gn}[1]{}
\renewcommand{\sz}[1]{}

\usepackage{microtype}

\aclfinalcopy 
\def\aclpaperid{2749} 

\title{Soft Gazetteers for Low-Resource Named Entity Recognition}

\author{Shruti Rijhwani, Shuyan Zhou, Graham Neubig, Jaime Carbonell \\
  Language Technologies Institute \\
  Carnegie Mellon University \\
  \texttt{\{srijhwan,shuyanzh,gneubig,jgc\}@cs.cmu.edu} \\}

\date{}

\begin{document}
\maketitle
\begin{abstract}
Traditional named entity recognition models use gazetteers (lists of entities) as features to improve performance. Although modern neural network models do not require such hand-crafted features for strong performance, recent work~\cite{wu-etal-2018-evaluating} has demonstrated their utility for named entity recognition on English data. However, designing such features for low-resource languages is challenging, because exhaustive entity gazetteers do not exist in these languages. To address this problem, we propose a method of ``soft gazetteers'' that incorporates ubiquitously available information from English knowledge bases, such as Wikipedia, into neural named entity recognition models through cross-lingual entity linking. Our experiments on four low-resource languages show an average improvement of 4 points in F1 score.\footnote{Code and data are available at \url{https://github.com/neulab/soft-gazetteers}.}

\end{abstract}

\section{Introduction}
\label{sec:intro}

Before the widespread adoption of neural networks for natural language processing tasks, named entity recognition (NER) systems used linguistic features based on lexical and syntactic knowledge to improve performance \cite{ratinov-roth-2009-design}.
With the introduction of the neural LSTM-CRF model~\cite{DBLP:journals/corr/HuangXY15,lample-etal-2016-neural}, the need to develop hand-crafted features to train strong NER models diminished. 
However, \citet{wu-etal-2018-evaluating} have recently demonstrated that integrating linguistic features based on part-of-speech tags, word shapes, and manually created lists of entities called \emph{gazetteers} into neural models leads to better NER on English data.
Of particular interest to this paper are the gazetteer-based features -- binary-valued features determined by whether or not an entity is present in the gazetteer.
 
Although neural NER models have been applied to low-resource settings~\cite{cotterell-duh-2017-low,huang-etal-2019-cross}, directly integrating gazetteer features into these models is difficult because gazetteers in these languages are either limited in coverage or completely absent. Expanding them is time-consuming and expensive, due to the lack of available annotators for low-resource languages~\cite{strassel-tracey-2016-lorelei}.

As an alternative, we introduce ``soft gazetteers", a method to create continuous-valued gazetteer features based on readily available data from high-resource languages and large English knowledge bases (e.g., Wikipedia). More specifically, we use entity linking methods to extract information from these resources and integrate it into the commonly-used CNN-LSTM-CRF NER model~\cite{ma-hovy-2016-end} using a carefully designed feature set. We use entity linking methods designed for low-resource languages, which require far fewer resources than traditional gazetteer features~\cite{upadhyay-etal-2018-joint,zhou20tacl}.

Our experiments demonstrate the effectiveness of our proposed soft gazetteer features, with an average improvement of 4 F1 points over the baseline, across four low-resource languages: Kinyarwanda, Oromo, Sinhala, and Tigrinya. 

\section{Background}
\label{sec:background}
\begin{figure*}[tb]
    \centering
    \includegraphics[width=\textwidth]{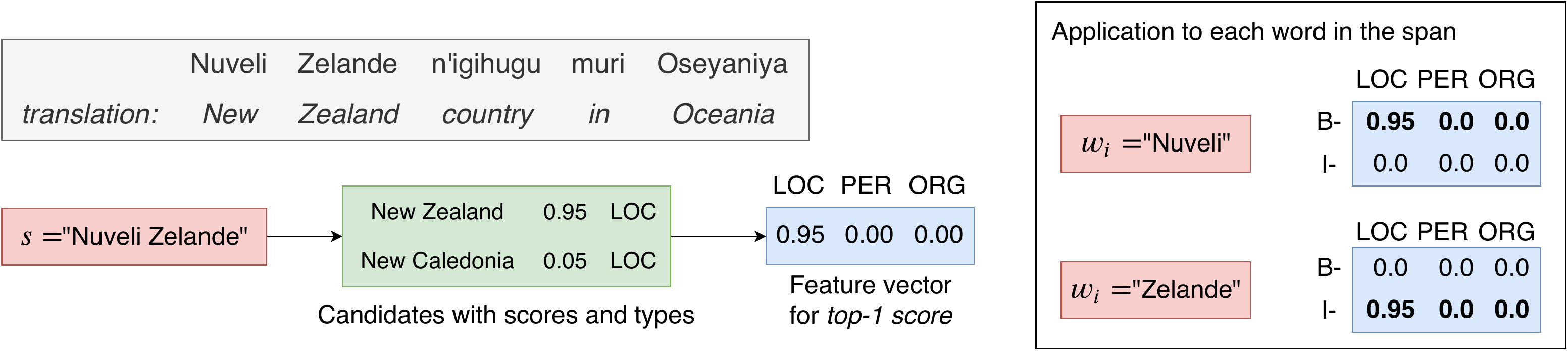}
    \caption{An example in Kinyarwanda to demonstrate soft gazetteer feature creation for each span $\boldsymbol{s}$ using candidate lists. The feature vector is applied to each word $\boldsymbol{w}_i$ in the span, depending on the position (``B-'' or ``I-'').
    }
    \label{fig:features}
\end{figure*}

\noindent
\textbf{Named Entity Recognition}\quad
NER identifies named entity spans in an input sentence, and classifies them into predefined types (e.g., location, person, organization).  
A commonly used method for doing so is the \textbf{BIO} tagging scheme, representing the \textbf{B}eginning, the \textbf{I}nside and the \textbf{O}utside of a text segment~\cite{ratinov-roth-2009-design}.
The first word of a named entity is tagged with a ``B-'', subsequent words in the entity are ``I-'', and non-entity words are ``O''. For example:

\smallskip
\noindent
[Mark]\textsubscript{B-PER} [Watney]\textsubscript{I-PER} [visited]\textsubscript{O} [Mars]\textsubscript{B-LOC}

\medskip
\noindent
\textbf{Binary Gazetteer Features}\quad Gazetteers are lists of named entities collected from various sources (e.g., nation-wide census, GeoNames, etc.). They have been used to create features for NER models, typically binary features indicating whether the corresponding $n$-gram is present in the gazetteer.

\medskip
\noindent
\textbf{Entity Linking}\quad
Entity linking (EL) is the task of associating a named entity mention with its corresponding entry in a structured knowledge base (KB)~\cite{hachey2013evaluating}. For example, linking the entity mention ``Mars'' with its Wikipedia entry.

In most entity linking systems~\cite{hachey2013evaluating,sil2018neural}, the first step is shortlisting candidate KB entries, which are further processed by an entity disambiguation algorithm. Candidate retrieval methods, in general, also score each candidate with respect to the input mention.

\section{Soft Gazetteer Features}
\label{sec:softgaz}

As briefly alluded to in the introduction, creating binary gazetteer features is challenging for low-resource languages. The soft gazetteer features we propose instead take advantage of existing limited gazetteers and English knowledge bases using low-resource EL methods. In contrast to typical binary gazetteer features, the soft gazetteer feature values are continuous, lying between 0 and 1.

Given an input sentence, we calculate the soft gazetteer features for each span of $n$ words, $\boldsymbol{s}=\boldsymbol{w}_i,\ldots,\boldsymbol{w}_{i+n-1}$, and then apply the features to each word in the span.
We assume that we have an EL candidate retrieval method that returns candidate KB entries $\mathcal{C}=(c_1, c_2...)$ for the input span.
$c_1$ is the highest scoring candidate.

As a concrete example, consider a feature that represents the \textit{score of the top-1 candidate}.
\autoref{fig:features} shows an example of calculating this feature on a sentence in Kinyarwanda, one of the languages used in our experiments. The feature vector $\boldsymbol{f}$ has an element corresponding to each named entity type in the KB (e.g.,~LOC, PER, and ORG).

For this feature, the element corresponding to the entity type of the highest scoring candidate $c_1$ is updated with the score of the candidate. That is,
$$\boldsymbol{f}_{\texttt{type}(c_1)}=\texttt{score}(c_1).$$
This feature vector is applied to each word in the span, considering the position of the specific word in the span according to the BIO scheme; we use the ``B-'' vector elements for the first word in the span, ``I-'' otherwise.

For a word $\boldsymbol{w}_i$, we combine features from different spans by performing an element-wise addition over vectors of all spans of length $n$ that contain $\boldsymbol{w}_i$.
The cumulative vector is then normalized by the number of spans of length $n$ that contain $\boldsymbol{w}_i$, so that all values lie between 0 and 1.
Finally, we concatenate the normalized vectors for each span length $n$ from 1 to $N$ ($N=3$ in this paper).

We experiment with different ways in which the candidate list can be used to produce feature vectors. The complete feature set is:

\begin{enumerate}
    \item \textbf{top-1 score}: This feature takes the score of the highest scoring candidate $c_1$ into account.
    $$\boldsymbol{f}_{\texttt{type}(c_1)} = \texttt{score}(c_1)$$
    \item \textbf{top-3 score}: Like the top-1 feature, we additionally create feature vectors for the second and third highest scoring candidates.
    \item \textbf{top-3 count}: These features are type-wise counts of the top-3 candidates. Instead of adding the score to the appropriate feature element, we add 1.0 to the current value. For a candidate type $t$,  such as LOC, PER or ORG,
    $$\boldsymbol{f}_t=\sum_{\mathclap{c \in \{c_1,c_2,c_3\}}} 1.0 \times \mathbf{1}_{\texttt{type}(c)=t}$$
    $\mathbf{1}_{\texttt{type}(c)=t}$ is an indicator function that returns 1.0 when the candidate type is the same as the feature element being updated, 0.0 otherwise.
    \item \textbf{top-30 count}: This feature computes type-wise counts for the top-30 candidates.
    \item \textbf{margin}: The margin between the scores of consecutive candidates within the top-4. These features are not computed type-wise. For example the feature value for the margin between the top-2 candidates is, 
    $$\boldsymbol{f}_{c_1,c_2}=\texttt{score}(c_1)-\texttt{score}(c_2)$$
\end{enumerate}{}

We experiment with different combinations of these features by concatenating their respective vectors. The concatenated vector is passed through a fully connected neural network layer with a $tanh$ non-linearity and then used in the NER model.

\section{Named Entity Recognition Model}
\label{sec:model}

\begin{figure}[tb]
    \centering
    \includegraphics[width=\columnwidth]{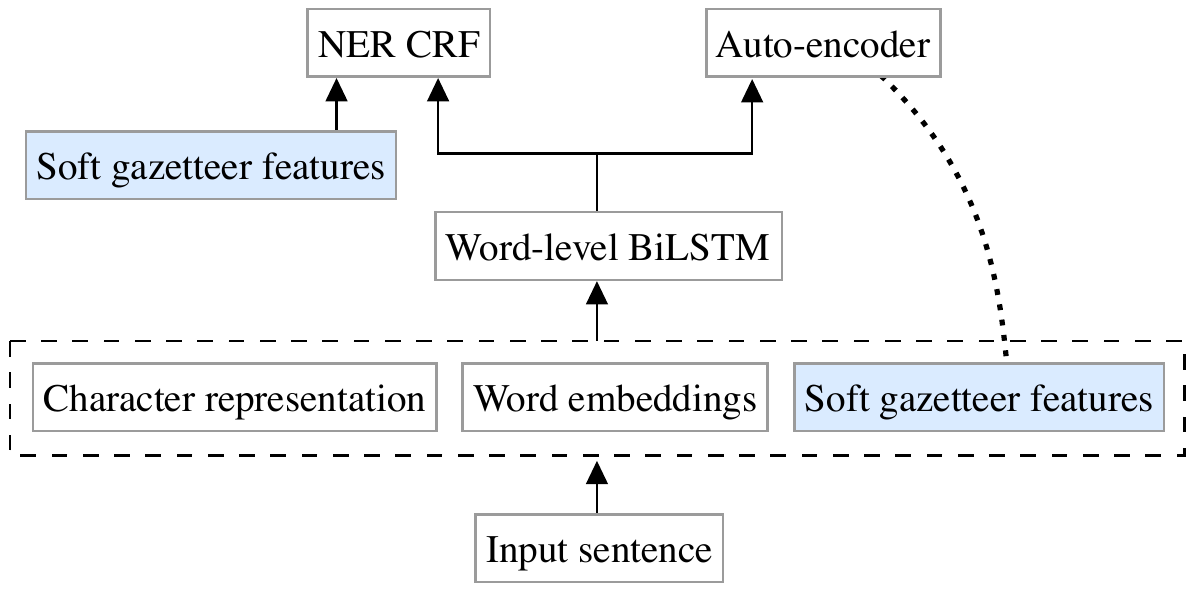}
    \caption{NER Model Architecture. The proposed soft gazetteer features are highlighted and the autoencoder reconstructs these features, indicated by a dotted line.
    }
    \label{fig:model}
\end{figure}

As our base model, we use the neural CRF model of~\citet{ma-hovy-2016-end}.
We adopt the method from~\citet{wu-etal-2018-evaluating} to incorporate linguistic features, which uses an autoencoder loss to help retain information from the hand-crafted features throughout the model 
(shown in \autoref{fig:model}).
We briefly discuss the model in this section, but encourage readers to refer to the original papers for a more detailed description.

\medskip
\noindent
\textbf{NER objective}\quad Given an input sequence, we first calculate a vector representation for each word by concatenating the character representation from a CNN, the word embedding, and the soft gazetteer features. The word representations are then used as input to a bidirectional LSTM (BiLSTM). The hidden states from the BiLSTM and the soft gazetteer features are input to a Conditional Random Field (CRF), which predicts a sequence of NER labels. The training objective, $\mathcal{L}_{CRF}$, is the negative log-likelihood of the gold label sequence.

\medskip
\noindent
\textbf{Autoencoder objective}\quad~\citet{wu-etal-2018-evaluating} demonstrate that adding an autoencoder to reconstruct the hand-crafted features leads to improvement in NER performance. The autoencoder takes the hidden states of the BiLSTM as input to a fully connected layer with a sigmoid activation function and reconstructs the features. This forces the BiLSTM to retain information from the features. The cross-entropy loss of the soft gazetteer feature reconstruction is the autoencoder objective, $\mathcal{L}_{AE}$.

\medskip
\noindent
\textbf{Training and inference} \quad The training objective is the joint loss: $\mathcal{L}_{CRF} + \mathcal{L}_{AE}$. The losses are given equal weight, as recommended in~\citet{wu-etal-2018-evaluating}.
During inference, we use Viterbi decoding to obtain the most likely label sequence.

\section{Experiments}
\label{sec:experiment}

In this section, we discuss our experiments on four low-resource languages and attempt to answer the following research questions:
1) ``Although gazetteer-based features have been proven useful for neural NER on English, is the same true in the low-resource setting?'' 2) ``Do the proposed soft-gazetteer features outperform the baseline?'' 3) ``What types of entity mentions benefit from soft gazetteers?" and 4) ``Does the knowledge base coverage affect performance?''.

\subsection{Experimental setup}
\begin{table}[tb]
    \centering
    \begin{tabular}{@{}lrrr@{}}
    \toprule
         Lang. & Dataset size & Frac. of NIL & Gaz. size \\
         \midrule
         kin & 951 & 0.41 & 912\\
         orm & 2958 & 0.36 & 313\\
         sin & 1068 & 0.29 & 2738\\
         tir & 2202 & 0.28 & 92\\
     \bottomrule
    \end{tabular}
    \caption{NER dataset and Wikipedia gazetteer sizes.}
    \label{tab:dataset}
\end{table}{}

\noindent
\textbf{NER Dataset}\quad We experiment on four low-resource languages: Kinyarwanda (kin), Oromo (orm), Sinhala (sin), and Tigrinya (tir). We use the LORELEI dataset~\cite{strassel-tracey-2016-lorelei}, which has text from various domains, including news and social media, 
annotated for the NER task.

\autoref{tab:dataset} shows the number of sentences annotated. The data is annotated with four named entity types: locations (LOC), persons (PER), organizations (ORG), and geopolitical entities (GPE). Following the CoNLL-2003 annotation standard, we merge the LOC and GPE types~\cite{tjong-kim-sang-de-meulder-2003-introduction}. Note that these datasets are very low-resource, merely 4\% to 13\% the size of the CoNLL-2003 English dataset. 

These sentences are also annotated with entity links to a knowledge base of 11 million entries, which we use \textit{only} to aid our analysis. Of particular interest are ``NIL" entity mentions that do not have a corresponding entry in the knowledge base~\cite{blissett-ji-2019-cross}. The fraction of mentions that are NIL is shown in \autoref{tab:dataset}.

\medskip
\noindent
\textbf{Gazetteer Data}\quad We also compare our method with binary gazetteer features, using entity lists from Wikipedia, the sizes of which are in \autoref{tab:dataset}.

\medskip
\noindent
\textbf{Implementation}\quad Our model is implemented using the DyNet toolkit~\cite{dynet}, and we use the same hyperparameters as~\citet{ma-hovy-2016-end}. We use randomly initialized word embeddings since we do not have pretrained vectors for low-resource languages.\footnote{A note on efficiency: our method involves computing entity linking candidates for each n-gram span in the dataset. The most computationally intensive candidate retrieval method (\textsc{Pbel}, discussed in \autoref{sec:methods}) takes $\approx$1.5 hours to process all spans on a single 1080Ti GPU. Note that this is a preprocessing step and once completed, it does not add any extra computational cost to the NER training process.}

\medskip
\noindent
\textbf{Evaluation}\quad We perform 10-fold cross-validation for all experiments because of the small size of our datasets. Our primary evaluation metric is span-level named entity F1 score.

\subsection{Methods}
\label{sec:methods}

\noindent
\textbf{Baselines}\quad We compare with two baselines:
\begin{itemize}
    \item \textsc{NoFeat}: The CNN-LSTM-CRF model (\autoref{sec:model}) without any features.
    \item \textsc{BinaryGaz}: We use Wikipedia entity lists (\autoref{tab:dataset}) to create binary gazetteer features.
\end{itemize}

\noindent
\textbf{Soft gazetteer methods}\quad We experiment with different candidate retrieval methods designed for low-resource languages. These are trained \textit{only} with small bilingual lexicons from Wikipedia, of similar size as the gazetteers (\autoref{tab:dataset}).
\begin{itemize}
    \item \textsc{WikiMen}: The WikiMention method is used in several state-of-the-art EL systems~\cite{sil2018neural, upadhyay-etal-2018-joint}, where bilingual Wikipedia links are used to retrieve the appropriate English KB candidates.
    \item Pivot-based-entity-linking \citep{zhou20tacl}: This method encodes entity mentions on the character level using n-gram neural embeddings~\cite{wieting2016charagram} and computes their similarity with KB entries. We experiment with two variants and follow \citet{zhou20tacl} for hyperparameter selection: 

    1) \textsc{PbelSupervised}: trained on the small number of bilingual Wikipedia links available in the target low-resource language. 
 
    2) \textsc{PbelZero}: trained on some high-resource language (``the pivot") and transferred to the target language in a zero-shot manner. The transfer languages we use are Swahili for Kinyarwanda, Indonesian for Oromo, Hindi for Sinhala, and Amharic for Tigrinya.
\end{itemize}

\noindent
\textbf{Oracles}\quad As an upper-bound on the accuracy, we compare to two artificially strong systems:
\begin{itemize}
    \item \textsc{OracleEL}: For soft gazetteers, we assume perfect candidate retrieval that always returns the correct KB entry as the top candidate if the mention is non-NIL.
    \item \textsc{OracleGaz}: We artificially inflate \textsc{BinaryGaz} by augmenting the gazetteer with all the named entities in our dataset.
\end{itemize}

\subsection{Results and Analysis}

Results are shown in \autoref{tab:results}.
First, comparing \textsc{BinaryGaz} to \textsc{NoFeat} shows that traditional gazetteer features help somewhat, but gains are minimal on languages with fewer available resources.\footnote{We note that binary gazetteer features usually refer to simply using the gazetteer as a lookup~\cite{ratinov-roth-2009-design}. However, we also attempt to use \textsc{WikiMen} and \textsc{Pbel} for retrieval, with scores converted to binary values at a threshold of 0.5. \textsc{BinaryGaz} in \autoref{tab:results} is the best F1 score among these methods--this turns out to be the string lookup for all four languages. This is expected because, for low-resource languages, the other candidate retrieval methods are less precise than their high-resource counterparts. Binary-valued features are not fine-grained enough to be robust to this.}
Further, we can see that the proposed soft gazetteer method is effective, some variant thereof achieving the best accuracy on all languages.

For the soft gazetteer method, \autoref{tab:results} shows the performance with the best performing features (which were determined on a validation set): \textbf{top-1} features for Kinyarwanda, Sinhala and Tigrinya, and \textbf{top-30} features for Oromo.
Although Sinhala (sin) has a relatively large gazetteer (\autoref{tab:dataset}), we observe that directly using the gazetteer as recommended in previous work with \textsc{BinaryGaz}, does not demonstrate strong performance. On the other hand, with the soft gazetteer method and our carefully designed features, \textsc{PbelSupervised} works well for Sinhala (sin) and improves the NER performance.
\textsc{PbelZero} is the best method for the other three languages, illustrating how our proposed features can be used to benefit NER by leveraging information from languages closely related to the target. The improvement for Oromo (orm) is minor, likely because of the limited cross-lingual links available for training \textsc{PbelSupervised} and the lack of suitable transfer languages for \textsc{PbelZero}~\cite{rijhwani2019zero}.

Finally, we find that both \textsc{OracleGaz} and \textsc{OracleEL} improve by a large margin over all non-oracle methods, indicating that there is substantial headroom to improve low-resource NER through either the development of gazetteer resources or the creation of more sophisticated EL methods.

\begin{table}[tb]
    \centering
    \resizebox{\columnwidth}{!}{
    \begin{tabular}{@{}lcrrrr@{}}
        \toprule
        Model && \multicolumn{1}{c}{kin} & \multicolumn{1}{c}{orm} & \multicolumn{1}{c}{sin} & \multicolumn{1}{c}{tir}\\
        \midrule
        \textsc{NoFeat}
        && 67.16 & 71.07 & 49.68 & 75.44\\
        \textsc{BinaryGaz} 
        && 69.05 & 71.24 & 54.08 & 75.84\\
        \midrule
        \textsc{WikiMen} 
        && 68.36 & 71.58 & 51.34 & 75.69\\
        \textsc{PbelSuper.} 
        && 68.94 & 71.61 & \textbf{60.95} & 76.49\\
        \textsc{PbelZero} 
        && \textbf{69.92} & \textbf{71.75} & 51.69 & \textbf{76.99}\\
        \midrule
        \textsc{OracleEL} 
        && 82.89 & 87.69 & 81.98 & 89.85\\
        \textsc{OracleGaz} 
        && 93.38 & 94.71 & 94.00 & 94.43\\
        \bottomrule
    \end{tabular}
    }
    \caption{10-fold cross-validation NER F1 score. 
    The best performing feature combination is shown here.
    Bold indicates the best non-oracle system.
    }
    \label{tab:results}
\end{table}

\begin{table}[tb]
    \centering
    \resizebox{\columnwidth}{!}{
    \begin{tabular}{@{}lcrrcrr@{}}
    \toprule
        && \multicolumn{2}{c}{Non-NIL Recall} && \multicolumn{2}{c}{Unseen Recall} \\
        \cmidrule{3-4}\cmidrule{6-7}
         Lang. && Baseline & SoftGaz && Baseline & SoftGaz \\
         \midrule
         kin && 66.5 & 73.3 && 35.4 & 43.9\\
         orm && 72.0 & 72.8 && 49.5 & 51.9\\
         sin && 57.3 & 69.8 && 20.3 & 35.3\\
         tir && 79.2 & 80.9 && 38.9 & 41.5\\
         \midrule
         Avg. && 68.7 & \textbf{74.2} && 36.0 & \textbf{43.1} \\
     \bottomrule
    \end{tabular}
    }
    \caption{Recall for non-NIL mentions and mentions unseen in the training data. SoftGaz represents the best soft gazetteer model as seen in \autoref{tab:results}.}
    \label{tab:recall}
\end{table}{}

\medskip 
\noindent
\textbf{How do soft-gazetteers help?}\quad We look at two types of named entity mentions in our dataset that we expect to benefit from the soft gazetteer features: 1) non-NIL mentions with entity links in the KB that can use EL candidate information, and 2) mentions unseen in the training data that have additional information from the features as compared to the baseline. \autoref{tab:recall} shows that the soft gazetteer features increase the recall for both types of mentions by several points.

\begin{table}[tb]
    \centering
    \resizebox{\columnwidth}{!}{
    \begin{tabular}{@{}lrrrrr@{}}
    \toprule
         && \multicolumn{1}{c}{kin} & \multicolumn{1}{c}{orm} & \multicolumn{1}{c}{sin} & \multicolumn{1}{c}{tir}  \\
        \midrule
        Orig. KB && 69.92 & 71.71 & 60.95 & 76.58 \\
        NIL augment && 76.28 & 76.50 & 70.87 & 83.07\\
    \bottomrule
    \end{tabular}
    }
    \caption{NER F1 score of the best performing soft gazetteer model with the original KB and with augmenting NIL-clustered entity mentions.}
    \label{tab:kb}
\end{table}{}

\medskip
\noindent
\textbf{Knowledge base coverage}\quad \autoref{tab:recall} indicates that the soft gazetteer features benefit those entity mentions that are present in the KB. However, our dataset has a significant number of NIL-clustered mentions (\autoref{tab:dataset}). The ability of our features to add information to NIL mentions is diminished because they do not have a correct candidate in the KB. To measure the effect of KB coverage, we augment the soft gazetteer features with \textsc{OracleGaz} features, applied \textit{only} to the NIL mentions. Large F1 increases in \autoref{tab:kb} indicate that higher KB coverage will likely make the soft gazetteer features more useful, and stresses the importance of developing KBs that cover all entities in the document.

\section{Conclusion}
We present a method to create features for low-resource NER and show its effectiveness on four low-resource languages. Possible future directions include using more sophisticated feature design and combinations of candidate retrieval methods.

\section*{Acknowledgements}
Shruti Rijhwani is supported by a Bloomberg Data Science Ph.D. Fellowship. Shuyan Zhou is supported by the DARPA Information Innovation Office (I2O) Low Resource Languages for Emergent Incidents (LORELEI) program under Contract No. HR0011-15-C0114. We also thank Samridhi Choudhary for help with the model implementation and Deepak Gopinath for feedback on the paper.

\bibliography{acl2020}
\bibliographystyle{acl_natbib}

\end{document}